\documentclass[journal]{IEEEtran}
\usepackage[utf8]{inputenc}
\usepackage{amsmath}  
\usepackage[capitalize,nameinlink]{cleveref}
\usepackage{graphicx}
\usepackage{xcolor}
\usepackage{url}
\usepackage[many]{tcolorbox}
\usepackage[normalem]{ulem}
\usepackage{nle-chicago-patch}
\bibpunct{(}{)}{,}{a}{}{;}
\usepackage{linguex}  
\makeatletter
\let\c@ExNo=\c@equation
\makeatother
\usepackage{gensymb} 
\usepackage{pgfplots} 
\pgfplotsset{compat=1.3}
\begin{document}
\title{High-dimensional distributed semantic spaces for utterances}
\author{Jussi Karlgren$^{1}$ and Pentti Kanerva$^{2}$ \\
$^{1}$Gavagai and KTH Royal Institute of Technology, Stockholm, Sweden \\
$^{2}$Redwood Center for Theoretical Neuroscience, UC Berkeley, CA, USA}

\maketitle

\begin{tcolorbox}[colback=red!10!white,
                     colframe=red!20!black,
                     title=\textsc{This paper has been published in the Journal of Natural Language Engineering},  
                     center, 
                     valign=top, 
                     halign=left,
                     before skip=0.8cm, 
                     after skip=1.2cm,
                     center title, 
                     width=3in]

     Karlgren, Jussi, and Pentti Kanerva. "High-dimensional distributed semantic spaces for utterances." Natural Language Engineering 25, no. 4 (2019): 503-517.

  \end{tcolorbox}

\begin{abstract}
High-dimensional distributed semantic spaces have proven useful and
effective for aggregating and processing visual, auditory, and lexical
information for many tasks related to human-generated data.

Human language makes use of a large and varying number of features,
lexical and constructional items as well as contextual and
discourse-specific data of various types, which all interact to
represent various aspects of communicative information. Some of these
features are mostly local and useful for the organisation of
e.g. argument structure of a predication; others are persistent over
the course of a discourse and necessary for achieving a reasonable
level of understanding of the content.

This paper describes a model for high-dimensional representation for
utterance and text level data including features such as constructions
or contextual data, based on a mathematically principled and
behaviourally plausible approach to representing linguistic
information. The implementation of the representation is a
straightforward extension of {\em Random Indexing} models previously used
for lexical linguistic items. The paper shows how the implemented
model is able to represent a broad range of linguistic features in a
common integral framework of fixed dimensionality, which is 
computationally
habitable, and which is suitable as a bridge between symbolic
representations such as dependency analysis and continuous
representations used {\it e.g.\ } in classifiers or further machine-learning
approaches.  This is achieved with operations on vectors that 
constitute a powerful computational algebra, accompanied with
an associative memory for the vectors. 

The paper provides a technical overview of the framework and a worked
through implemented example of how it can be applied to various types of linguistic features. 

\end{abstract}

\section{Human language}
Human language is a general-purpose representation of human knowledge, and models to process it vary as to the degree they are bound to some task or some specific usage. The current trend in computational language representations is to train a model to represent regularities and patterns with as little explicit knowledge-based linguistic processing as possible, and recent advances in such general models for end-to-end learning to address linguistics tasks have been quite successful. Most of those approaches make little use of information beyond the occurrence or co-occurrence of words in the linguistic signal and take the single word to be the atomic unit. The framework proposed by us in this paper shows how high-dimensional models similar to the ones currently used as a standard processing model for word-level similarities can be extended to accommodate linguistic items and feature beyond lexical items, in a transparent and handy representation similar to what is currently the standard processing model for word-level information. We expect that this model can be used as a front end for further processing by machine learning approaches that expected their input to be in continuous form. 

\subsection{Requirements for a representation}
There are some basic qualities we want a representation to hold to. A representation should have descriptive and explanatory power, be practical and convenient for further application, be reasonably true to human performance, provide defaults to smooth over situations where a language processing component lacks knowledge or data, and provide constraints where the decision space is too broad.

{\em Neurophysiological plausibility} We want the model to be {\em non-compiling}, i.e. not need a separate step to accommodate a new batch of data and be available on-line during training. We want the model to exhibit {\em bounded growth}, meaning it should not to grow too rapidly with new data.

{\em Behavioural adequacy}
We want the model to be {\em incremental}, i.e. to improve its performance (however we choose to measure and evaluate performance) progressively with incoming data. While we want our model to rely on the surface form of the input, we do not acknowledge the necessity to limit the input analysis to be white-space based tokenisation: a more sophisticated model based on the identification of patterns or constructions in the input is as plausible as a naive one. We want our representation to allow for explicit inclusion of analysis results beyond the word-by-word sequences typically used as input to today's models. 

{\em Computational habitability}
We want the model to be evaluable and transparent, and manageable computationally in face of large and growing amounts of input data it is exposed to. We do not want it to make assumptions of a finite inventory of lexical items or expressions. We want the model to accommodate potential features of interest during experimentation without requiring expensive reconfiguration of the processing scheme. 

{\em Explicit representation of features}
We want the model to allow exploration by explicit inclusion of linguistic features of potential interest---This means that we want the model to accept input preprocessed to include more advanced feature sets than mere word tokens. 

{\em Context and Anchoring}
We want the model allow the inclusion of extra-linguistic data and annotations. Linguistic data is now available in new configurations, collected from situations which allow the explicit capture of location, time, participants, and other sensory data such as biometric data, meteorological data, and social context of the author or speaker. These data are potentially of great interest e.g. to resolve ambiguities or to understand anaphor and deictic reference and we see no reason for them to be represented separately from the linguistic signal.

\section{Linguistic theory}
The model we propose is built on two foundational theoretical approaches in linguistics. Common to most learning models our model builds on distributional semantics. Such models are typically built to track individual words and occasionally multi-word terms. We extend the model by viewing the linguistic signal from a radical construction grammar perspective, adding constructional items to the feature set to be observed. This allows us to model the information in an utterance with more detail than the constituent terms in it. 

However, it remains to be established what the proper purpose of an utterance representation would be. Lexical semantics can well be argued to be determined both by how a lexical item combines in a sequence with other items and what other items could be used validly in its place. For utterances, typical representations concern what linguistic items it is composed of and how they are combined, without paying much attention to how those combinations constrain the external context of the utterance. We would expect that a future framework for full representation of an utterance allows the representation of such contextual information in some form, whether it be discourse level models or even extra-linguistic information.

\subsection{Distributional semantics}
Distributional semantics is based on well-established philosophical and linguistic principles, most clearly formulated by Zellig Harris (\citeyear{harris1968mathematicalstructure}). Distributional semantic models aggregate observations of items in linguistic data and infer semantic similarity between linguistic items based on the similarity of their observed distributions. The idea is that if linguistic items---such as the words {\em herring} and {\em cheese}---tend to occur in the same contexts---say, in the vicinity of the word {\em smörgåsbord}---then we can assume that they have related meanings. This is known as the {\em distributional hypothesis} \citep{sahlgren2008distributional}.
Distributional methods have gained tremendous interest in the past decades, due to the proliferation of large text streams and new data-oriented learning computational paradigms which are able to process large amounts of data. So far distributional methods have mostly been used for lexical tasks, and include fairly little of more sophisticated processing as input. This is, to a great extent, a consequence of the simple and attractively transparent representations used. This paper proposes a model to accommodate both simple and more complex linguistic items within the same representational framework. 

Common to all distributional approaches is that they collect observations and generalise from them, whether they end up formulating their results as a probabilistic or a geometric model. The parameters of those approaches are manifold and impinge on the results in various ways. \cite{levy2015improving} show how those parameters are largely translatable between processing models and that the exact choice of processing model is in most respects less important than the choice of what information to process. Our claim here is that is desirable that the processing model and representation chosen adhere to the principles given above.

\subsection{Construction grammar}
The {\em Construction grammar} framework in its most radical formulations is characterised by the central claim that {\em lexical items}---the words---and their {\em configurations}---the syntax---are processed similarly or even identically. In construction grammar, lexical and syntactic observations are both viewed as linguistic items with equal salience and presence in the linguistic signal. This notion, that lexicon and syntax should be processed similarly, is referred to as the {\em syntax-lexicon continuum}. This is to be understood in contrast with most linguistic theories that divide the structural analysis of syntax and lexicon into separate representations and processing models \citep{croft2005radical}.
The parsimonious character of construction grammar is attractive as a framework for integrating a dynamic and learning view of language use with formal expression of language structure: it allows the representation of words together with constructions in a common framework, and makes no claims to what elements of constructions are obligatory, universal, or foundational. 

For our purposes construction grammar provides an elegant theoretical foundation for a consolidated representation of both individual items in utterances and their configuration.

\section {High-dimensional representation}

\subsection {General properties}\label{sec:genprop}
Vectors with thousands of dimensions have properties appropriate for
modeling cognitive processes. For example, if one fourth of the components have
changed or are corrupted by noise, the altered vector can still be
identified with the original---it is more similar to the original than
a randomly chosen vector would be.  This corresponds to our ability to
recognize faces and voices, identify animals and plants and so forth
with apparent ease, although the information available to the brain
never repeats exactly.  But the apparent ease is deceptive.  Trying to
build such traits into artificial systems has proven to
be very difficult.

This is true also of language.  Although words are discrete units,
they often have a range of meanings or several disjoint meanings.  Properties of high-dimensional
vectors are useful here as well.  If we represent meaning by a
high-dimensional vector, called a {\em semantic vector}, the vector for {\it
train} can at once be similar to vectors for {\it bus} and {\it teach}
(and for {\it coach!}) and dissimilar to nearly every other semantic
vector.  The apparent ease with which our brains deal with ambiguity
in language is very likely due to high-dimensional representation.

A model built from semantic vectors is a vector space which allows geometrical computational approaches to access the information collected into the space. Vector space models are frequently used in practical information access, both for research experiments and as a building block for systems in practical use at least since the early 1970's \citep{salton1975vectorspace, dubin2004vectorspacesurvey}. 

Vector space models have attractive qualities: processing vector spaces is a manageable implementational framework, they are mathematically well-defined and understood, and they are intuitively appealing, conforming to everyday metaphors such as ``near in meaning'' \citep{schutze1993wordspace}.

Semantic vectors have attractive qualities for implementations of distributional semantic theories. The content of a semantic vector can be accrued from observing contexts in which the word has been observed. If those contexts also are represented in some vector form, the aggregation of observations into a semantic context vector is straightforward. The selection of the most appropriate context defines what semantic relations are represented in the model \citep{sahlgren2006wordspace}.

The vector space model for meaning is the basis for most all information retrieval experimentation and implementation, most machine learning experiments, and is now the standard approach in most categorisation schemes, topic models, deep learning models, and other similar approaches, including this present model. 

Common to semantic vector spaces is that the dimensionality of a vector space grows rapidly with $N$, the number of items (words, in practice) and $C$, the number of contexts observed. Most implemented models rely on some form of dimensionality reduction, which allows the large and sparse raw observations to be reduced into a denser and more manageable form. Latent Semantic Analysis or Indexing (LSA or LSI, respectively) is among the earliest and better known methods \citep{deerwester1990indexing,landauer1997solution}.  Latent Semantic Indexing in its original implementation makes note of what documents a word has occurred in and reduces the observations of words by contexts using singular-value decomposition to build several-hundred-dimensional semantic vectors for words.  Semantic vectors for phrases, sentences, paragraphs, news articles, and so forth are then computed by summing over vectors for the words. Recent distributional models such as the neurally inspired deep learning models avoid the explicit dimensionality reduction step by training a transition matrix that transforms a sparse incoming vector of dimensionality $N$ into some manageable internal processing dimensionality, typically of $o(100)$. Training this matrix typically is a major computational effort.

The method of {\em random projections} is an efficient alternative to dimensionality reduction steps that takes advantage of high dimensionality \citep{papadimitriou2000latent}. {\em Random Indexing}  is a simple implementation of random projections for building semantic vectors, and provides us with an apt introduction to
computing with high-dimensional vectors \citep{kanerva2000random,sahlgren2008permutations}.

In previous experiments for learning lexical similarity between words, 
we have used random indexing with 2,000-dimensional vectors.  Each
word in the vocabulary is represented by {\it two} such vectors, one
constant and the other variable.  The constant vector is called the
word's index vector or {\it label} and is assigned to the word at random when the
word is first encountered.  Sparse ternary labels have worked well: a
small number of $+1$s and the same number of $-1$s (e.g., 10 each)
randomly placed among 1,980 zeros.  The variable vector is the word's
semantic vector.  It starts out as the zero-vector and is updated
every time the word occurs in the corpus.

The text is read by focusing on one word at a time, and a few words
before and after the {\it focus word} are its {\it context window}.
The semantic vector for the focus word is then updated by the random
labels of the context words.  If the text under consideration is as in Example \ref{fox} at the point when the
reading has progressed to the word {\it over} the
semantic vector for {\it over} will be updated by adding to it the random
labels for its context (marked by brackets in the example) {\it brown}, {\it fox}, {\it jumped}, {\it the}, {\it lazy}
and {\it dog's}. Thus, a word's semantic vector will be the {\it sum} of the random labels
of its observed neighbours. In this example the context from which the semantic vector is learned is a symmetric window 3 + 3 words wide. How that context is chosen and encoded---what its range is, whether words should be weighted according to distance from the focus word or their global occurrence statistics, whether some words should be excluded from the calculation entirely, whether left-hand neighbours should be recorded separately from right-hand neighbours---has great effects on the resulting semantic model \citep{levy2015improving}. 

\ex.\label{fox}
\a. The quick [brown fox jumped] {\bf over} [the lazy dog's] back.

A major issue with early experiments with semantic vectors for utterance-level analysis was the absence of
linguistic structure in the analysis. Proximity of words in the corpus
was all that was taken to matter, and so first examples of vectors for utterances represented the topic being
discussed but leave out the story being told---whether it was the boy
who hit the ball or the ball that hit the boy. Later models introduce representations to 
allow parts of speech and constituent structure to be encoded into the
vectors \citep{baroni2010nouns, wu2011structured}, and in have found that relatively simple manipulation of vectors through addition and multiplication can yield useful representation of some phrasal structures \citep{mitchell2008vector, mitchell2010composition}. The at time of writing  most broadly established neurally inspired model, word2vec, retains some structural information in its semantic space and this can be used to infer some phrasal structure \citep{mikolov2013distributed}. Several efforts to include more information from linguistic analyses have been done \citep[e.g.]{pado2007dependency,baroni2010distributional} but these efforts have typically foundered on sparsity of data and impracticality of the linguistic preprocessing in face of non-standard text. Most recent efforts have discounted the necessity of working with parsing or other intra-sentential structural information, trusting in the ability of end-to-end models which are taught to tailor the analysis to some task to generalise. While the performance on current tasks for such models is impressive, we expect further tasks to again motivate the use and introduction of structural information in utterance representations. There are many potential approaches to take, and while most are based on dependency formalisms \citep[e.g.]{weeds2014distributional}, we also find with some interest that some linguistic structures in compositional models of distributional semantics have been formulated in terms of categorial grammar \citep{clark2016categorial}. The inclusion of such information, whether through phrase structure models or dependency models will by necessity entail more complex representations and current work attempts to manage this through e.g. tensor models which allow a confluence of information to be represented simultaneously. These models come at a considerable computational and conceptual cost \citep[e.g.]{polajnar2014reducing, sandin2017random}. The approach we present here allows for explict experimentation, e.g. by including dependency information along with lexical items, allowing for concurrent application of many information sources in one representation, but does so in a computationally and conceptually habitable manner, obviating the need for tensor models.

\subsection{Hyperdimensional computing}

In this paper we describe simple operations on high-dimensional
vectors that allow both bag-of-words semantics and constituent
structure to be expressed in the same vector. We make use of the framework first
introduced by Plate under the name Holographic Reduced Representation
(HRR; \citealp{plate1991holographic,plate2003holographic}). Its variants go by different names: MAP (for
Multiply--Add--Permute; \citealp{gayler1998multiplicative}), Vector Symbolic Architecture
(VSA; \citealp{gayler2004vector}), and Hyperdimensional Computing \citep{kanerva2009hyperdimensional}.
These systems encode information with {\it three} operations that keep
vector {\it dimensionality constant,} thereby allowing composition of
structure to any depth.  Two of the
operations correspond to addition and multiplication, and the third
permutes or shuffles vector coordinates, as detailed below in Section~\ref{sec:mathdef}.  A key property to the
operations is that they allow fully general computing based on a
well-understood computational algebra.  The details of addition and
multiplication depend on the kinds of vectors used, whether binary,
bipolar, ternary, integer, real or complex. This framework---a vector space 
together with linear algebraic manipulation operations and geometric access and analysis operations---can 
be used for the purposes of a richer linguistic representation which allows us to represent utterances, including elements of their structure, in a common vector representation. The training model used by word2vec and related neural approaches has previously been combined with random indexing and linear algebraic operations to encode certain syntactic relations in texts from the biomedical domain to establish similarities between concepts \citep{widdows2014reasoning, cohen2017embedding}. The framework presented here is quite closely related to that work, but aims to generalise it to a larger potential inventory of features.


For the purposes of the following discussion, we will make use of $n$-dimensional vectors populated with both positive and negative real values, with $n > 1,000$. Computing with high-dimensional vectors begins by assigning randomly generated index vectors or labels to basic observed items, with independent, identically distributed components. In a study of orthography, letters would be basic objects, and in a study of lexical semantics, words or multi-word terms would be. In this present model, constructional linguistic units are included as basic objects along with lexical items by assigning them labels along with everything else of interest. Starting with the index vectors we compute vectors for composed entities using the three operations given above.  For example, if $\bar{v}_m$, $\bar{v}_a$ and $\bar{v}_p$ are vectors for the letters {\it m}, {\it a} and {\it p}, the sum vector $\bar{v}_{map} = \bar{v}_m + \bar{v}_a + \bar{v}_p$ is a bag-of-letters representation of the word {\it map}: it is similar to each of the vectors $\bar{v}_m$, $\bar{v}_a$ and $\bar{v}_p$, and it the same as the vectors for {\it amp} and {\it pam}.  However, if we want to have a vector that is unique to {\it map} and dissimilar to all other word vectors, we can make use of permutation operations or multiplication operations to distinguish the positions of  the vectors $\bar{v}_m$, $\bar{v}_a$ and $\bar{v}_p$ in the sequence. 
Sequential structure can be encoded in various ways and we will return to this issue in the implemented example in the next section.

{\it Similarity} ($\sim$) of vectors is calculated based on the angle between them or the distance between them on the unit sphere. Scalar or dot product can be used as such or normalized as cosine or Pearson correlation.  Cosine = 1 means most similar and cosine = 0
means dissimilar, unrelated, orthogonal. Through the geometric properties of high-dimensional
spaces, any given vector will be quite dissimilar to any other randomly picked vector unless some of the information in them has caused them to converge.  What cosine values are interesting or notable must be calibrated for each implementation, depending on amounts of data, dimensionality, and density of representation; in general, a cosine or correlation
of about 0.25 or better between two vectors will mean that they are notably similar.

\subsection {Mathematical specifics}\label{sec:mathdef}

The following is an overview of the most important properties of the
operations, with examples of their use in encoding and decoding
information.  

An {\em addition} of vectors, resulting in a sum vector is {\it similar} to its operand vectors
($A + B \sim A$) and independent of their order ($A + B = B + A$); it
can be used to represent a set or a multiset (``bag'' is another name for
multiset, hence ``bag-of-words'').  The similarity between the sum and
its operands decreases with the number of vectors in the sum.  Two sum
vectors are similar if most of their operands are the same, e.g.,
\begin {align*}
          & A + B + \ldots + T + U + V \\
  \sim \  & A + B + \ldots + T + X + Y + Z
\end {align*}
This is the idea behind semantic vectors as bags of words.

{\it Multiplication} is done {\it coordinate by coordinate}, known as the Hadamard product.
Multiplication is {\it invertible} and the product vector is
{\it dissimilar} to its operands ($A*B \not\sim A$). In particular, a bipolar vector---a vector populated with $1$s and $-1$s multiplied by itself is a vector of 1s, meaning that the vector is its own inverse. Multiplication can be used
for variable binding for any variable of interest. For our purposes here, this could be grammatical features related to e.g. agreement resolution, such as number or tense, or situational data such as time of day, number of speaker, or any other item of interest under study. Inverse multiplication can then be used to
release the value bound to a variable.  For example, if we assign a vector $X$ to represent a variable $x$ and another variable $A$ for a value $a$ which that variable can take, we can use $X*A$ 
to represent the fact that variable $x$ has value $a$.  We can then, if $X$ has been defined to be a bipolar vector, 
recover the value by multiplying that product again with $X$:
$$
   X*(X*A) = (X*X)*A = {\mathbf 1}*A = A
$$

Multiplication {\it distributes} over addition: $X*(A + B) = (X*A) +
(X*B)$.  This makes it possible to add several bound variables into a
single vector and to recover bound values.  For example, we can encode
the values of three variables $\{x = a, y = b, z = c\}$ using bipolar vectors $X$, $Y$, and $Z$ added into a 
vector $(X*A) + (Y*B) + (Z*C)$ and then find the value that is bound
to $X$ by multiplying with $X$ as above:
\begin {align*}
   X *((X*A) + (Y*B) + (Z*C)) & = X*(X*A) + X*(Y*B) + X*(Z*C)) \\
      & = A \qquad +\quad \mathrm{noise} \quad + \quad \mathrm{noise} \\
      & \sim A 
\end {align*}
The answer is approximate but close enough to the exact vector to be
identified with it with very high probability.  However, as the number
of bound pairs in the sum vector increases, the ability to recover
values of bound variables decreases \citep{frady2018theory}.

Multiplication {\it preserves similarity}.  If two vectors are
multiplied by the same third vector, the resulting vectors are just as
similar to each other as the originals:
$$
  \mathrm{sim}(X*A, X*B) = \mathrm{sim}(A, B)
$$
This suggests a mechanism for computing with analogy.

{\it Permutation} takes a single operand, rearranges its coordinates,
and produces a vector that is {\it dissimilar} to the operand ($\Pi(A)
\not\sim A$).  Permutations resemble multiplication in several ways:
they are {\it invertible}, {\it preserve similarity}, and {\it
  distribute} over addition.  They distribute also over multiplication
($\Pi(A*B) = \Pi(A) * \Pi(B)$), making them extremely useful for
encoding and decoding compositional structure.  However, permutations
are matrices rather than vectors and so they are not elements of the
space of representations.  In mathematical terms, they are unary
operations on vectors whereas multiplication is a binary operation.
In practice this means that a vector used for multiplication can be
learned within the system, whereas permutations must be predefined.
They can, however, be operated on by other permutations. The number of possible permutations is enormous.

Examples of (multi)sets, sequences, and variable binding were shown
above.

Here we present one more, the encoding of nested structures.
If we encode the pair $(a,b)$ with two unrelated
permutations $\Pi_{car}$ and $\Pi_{cdr}$ as $\Pi_{car}(A) + \Pi_{cdr}(B)$ then the
nested structure $((a, b),(c, d))$ can be represented by
 \begin {equation}
 \begin{split}
   & \Pi_{car}(\Pi_{car}(A) + \Pi_2(B)) + \Pi_2(\Pi_{car}(C) + \Pi_{cdr}(D)) \\
   & = \Pi_{carcar}(A) + \Pi_{carcdr}(B) + \Pi_{cdrcar}(C) + \Pi_{cdrcdr}(C)
\end{split}
\end {equation}
where $\Pi_{ij}$ is the permutation $\Pi_i \Pi_j$.

The ability to decode distributed representation and to recover its
constituents is essential to computing with high-dimensional vectors.
There is of course a limit to the amount of information that can be
represented in a single vector, and it depends on dimensionality and
on the value range of individual components.  Generally speaking, the
ability to resolve a sum vector into its constituent vectors grows
linearly with dimensionality, and the ability to determine whether a
given vector is included in a sum vector grows exponentially with
dimensionality \citep{gallant2013representing,frady2018theory}.


The above overview introduces the operations used in the rest of the
paper.  There are deeper reasons for having discussed them in as much
detail as we have done here.  Making semantic vectors with random
indexing mimics how we learn language and assign meaning to words.
The random index vectors are like words: they are distinct and constant
and get their meaning from their use with other words.  The fact that
the words of a language are rather arbitrary agrees with the notion
that they are in essence random, yet can serve as the material on
which meaning relations---the semantic vectors---are built.

There is a further reason for discussing the mathematics in detail.
To come to terms with the complexity and fluidity of language, we need
a rich and powerful system of representation, the workings of which
can also be understood.  That is the reason for drawing attention to
(quasi)orthogonality among high-dimensional vectors and invertibility
and distributivity of operations on them.  Computing in distributed
representation with high-dimensional random vectors based on their
algebra is a candidate for such a system.  Our task is to find out how
it maps to various aspects of language.

\section{An implemented example}
The aim of our framework is to be able to represent an utterance---a clause or a sentence or some similar chunk of language. Previously, high-dimensional models have been used for the representation of lexical similarity, and while some of those results carry over to utterance models there are specifics to attend to when extending the model to utterances. The objective is to generate for each utterance a vector which represents the distinctive features of that utterance. Those features will include what lexical items the utterance is composed of which is something an additive combination of referential expressions will handle with some ease, but we wish to be able to include both observable constructional items as well as semantic roles that the entities mentioned in the utterance participate in. In this implemented version we do not make use of the full potential expressiveness of the high-dimensional representation, nor of the potential feature space human language and its usage allows. 
There are many potential features which may be of interest, both linguistic (e.g. stylistic analyses) and extra-linguistic (e.g discourse participants, location, time of day) and introducing the possibility to operate experimentally with such features is one of the major design motivators of this present framework. Processing the linguistic signal and whatever extra-linguistic information is available to identify such features may be variously difficult, but the framework described here makes their inclusion in further processing straightforward: each feature of interest can be afforded its separate randomly generated label and included in the utterance vector through addition, multiplication, or permutation. The amount of information that fits into a high-dimensional vector is quite large and adding further features will add somewhat to noise, but not dramatically decrease the resolution of information already recorded in the vector. One of the fortes of this type of representation is that adding more information to a vector does not change its dimensionality which means that the processing pipeline need not be reconfigured if a new feature is added to the data under consideration.


\subsection{Data set}
We have recently used this type of representation for the study of author characteristics, 
question categorisation \citep{karlgrenkanerva2018hyperdimensional}, and language identification \citep{joshi2016language}. To demonstrate some of the versatility of this approach, we process a set of about 1 million microblog posts from Twitter and show how some various features can be encoded in vector form.\footnote{The posts were collected during the Fall of 2017, during which time period hurricanes Irma and Harvey caused damage and distress for much of the Caribbean and Southwestern United States. These posts were collected as part of a separate project on citizen observatories and public sentiment with respect to natural events, specifically flooding, and will be used in that project to investigate how attitude in writing co-varies with tense, mood, and aspect over an event timeline.} Some utterances from the data set are given in Example~\ref{sampleprobes}.

\ex. \label{sampleprobes}
 \a. \label{samplesamplea} Getting as far away from this hurricane as possible.
 \b. \label{samplesampleb} afraid
 \c. \label{samplesamplec} I am afraid of the hurricane
 \d. \label{samplesampled} I said I am afraid of the hurricane

\subsection{Vocabulary}
Each lexical item observed in the material is assigned an individual random vector as an index key. These are aggregated for each utterance by addition. Some terms contribute more to the distinctiveness of an utterance than others, and this can here, as in other similar quantitative models be accomplished through judicious weighting of terms, e.g. using {\sc TF-IDF} or {\sc ppmi} scoring. If the collection under consideration is static there are several well-established candidate approaches of understanding the relative informational and discriminative power of individual terms; if the data are streaming an online weighting scheme which can accommodate to changing statistics is more useful. In this example we will use a streaming weighting scheme shown in Equation \ref{eq:weight} developed for a large-scale lexical learning model \citep{sahlgren2016lexicon}.

\begin{equation}\label{eq:weight}
w(l) = e^{-{\lambda \cdot \frac{f(l)}{V}}}
\end{equation}

\noindent
In Equation \ref{eq:weight}, $w(l)$ is the weight of a linguistic item $l$, $\lambda$ is an integer that controls the aggressiveness of the frequency weight, $f(l)$ is the observed frequency of the item, and $V$ is the current size of the growing vocabulary or feature palette observed so far. This weighting formula returns a weight that ranges between close to 0 for very frequent terms, and close to 1 for less frequent terms. These are then used when adding lexical entries from an utterance to form a sum of vectors for the lexical content. 

\begin{equation}\label{eq:lexsum}
  \vec{U}_{lex} = \sum_{l \in \mathrm{utterance}} w(l) \times \bar{v}_l
\end{equation}

\noindent
As an example, the vector for the sentence in Example \ref{samplesamplea} will be a weighted sum of the nine constituent words' vectors; presumably, the words ``as'', ``this'', and ``from'' will have a rather low weight; other words higher. 

This implementation takes the simplest possible approach to lexical features, with no morphological normalisation and without training specific context vectors or using pre-trained models: at this point in the procedure, instead of using randomly generated index keys, previously trained context vectors from other lexical resources could be used in their place. This would provide a generalisation from representing tokens to representing concepts, if the context vectors in question were trained appropriately. 

Table~\ref{lexicalneighbours} shows which items are found to be closest neighbours to the probe sentences given in Example \ref{sampleprobes} by lexical items alone.  

\begin{figure}[h]\label{lexicalneighbours}
\begin{tabular}{c|cl}
\hline
Sample & cosine & utterance \\
\hline
  \ref{samplesamplea}   &  0.15   & I'm far away from the hurricane.  \\
  \ref{samplesampleb}   &  0.16  &  Lenny Bruce is not afraid. \\
  \ref{samplesamplec}   &  0.19   & i will always be afraid of hurricanes   \\
  \ref{samplesampled}   &  0.18   & like i said, i'm chillin through this hurricane.  \\
\end{tabular}
\caption{Closest neighbours to probe sentences in Example \ref{sampleprobes} by lexical measures}
\end{figure}

\subsection{Constructional Items}
The primary approach to include constructional linguistic items in the representation of utterances is to assign index vectors to them and then process those constructional items exactly as if they were lexical items. 

Some constructional items are observable without much specific analysis: an utterance can be negated, can be in past tense, can be qualified by some coherent or interesting set of adverbials. Some items involve non-trivial processing; others are closely bound to some set of lexical items. Any such observable item can be assigned a random index vector similar to those assigned to individual lexical items. In this implementation Example \ref{samplesamplec}, e.g. is assigned features ``present tense", ``first person singular pronoun subject", ``expression of fear and worry". Such features are easy to add and evaluate (naturally subject to the requirement they be observable with any accuracy). Some of the features added such as ``main verb appears late in clause" proved to have little utility for the analysis; others had better explanatory and discriminatory power. How constructional features should be weighted needs more thought: in our present implementation they have not been frequency weighted and are added in using Equation \ref{eq:lexsum} with the weights $w(l)$ set to 1 by default.

\subsection{Semantic roles}
Semantic roles are relations between the state or process and utterance refers to and entities referred to in the utterance. We encode a lexical item in a role by taking the index vector of the lexical item and permuting it by a permutation specific to its role, randomly generated. The utterance vector will then accommodate the lexical items as they occur, as a mention, and then again, permuted by semantic role. In our current implemented example we use dependency graphs from the Stanford CoreNLP \citep{manning2014stanfordcorenlp} to pick out the main verb from the main clause, the subject of that verb (if present and identified), and clausal and verbal adverbials.  They are encoded by taking the index vector $\bar{v}_r$ of the head word $r$ of the constituent in question and permuting it by a role-specific permutation $\Pi_s$ for role $s$. These are added to the lexical vector $\bar{U}_{lex}$ given by Equation \ref{eq:lexsum} above as shown in Equation \ref{eq:semrolesum}, to yield a resulting vector $\bar{U}_{roles}$.

\begin{equation}\label{eq:semrolesum}
  \bar{U}_{roles} = \bar{U}_{lex} + \sum_{r\ with\ role\ s} \Pi_{s}( \bar{v}_r )
\end{equation}

The vectors $\bar{v}_r$ are by default the same lexical vectors which were used for lexical models in Equation \ref{eq:lexsum}. Since they are permuted by $\Pi_s$ they will be practically orthogonal to the lexical vector. 

\subsection{Sequential structure}
The sequential structure of a sentence can be represented as a sequence of generalised labels for each token: most models of syntax can be translated into a label sequence, especially if bracketing labels are allowed. In this implementated example, in place of full dependency trees, we use triples of part of speech labels: more elaborate syntactic representations can be included similarly. Obviously, triples are a poor model of general syntactic structure: if this example is used as a model, the length of the subsequence and nesting levels should be determined by an informed assessment of what a typical constituent length in the syntactic model is and what generalisation one believes are valid to make from the data at hand. We take each utterance and label its words using the Natural Language Toolkit part of speech labeling with labels conforming to the Penn Treebank label set \citep{bird2009natural}. From the resulting sequence of lexical category labels all subsequences of length three are extracted. Example \ref{postag} shows how a sentence is converted to a set of overlapping triples.

A sequence of labels such as these triples can be represented in various ways. One way might be to assign an index vector for each symbol in the label palette and then to introduce a permutation for sequential relationships. Recalling the example {\em map} from Section~\ref{sec:genprop} above, we might want to represent it as a sequence distinct from {\em amp} and {\em pam}. One way to do so would be to define a permutation $\Pi_{precede}$ to represent the relative position of items, and then represent {\em map} through repeated application of $\Pi_{precede}$ to the character vectors $\bar{v}_m$, $\bar{v}_a$ and $\bar{v}_p$ and multiply the resulting vectors instead of addition: instead of the bag-of-letters sum vector $\bar{v}_{map} = \bar{v}_m + \bar{v}_a + \bar{v}_p$ we would have $\bar{v}_{map}$ = $\Pi_{precede}(\Pi_{precede}(\bar{v}_m)) * \Pi_{precede}(\bar{v}_a) * \bar{v}_p $. 

\ex. \label{postag}
\a. Anyone have a travel rest pillow I could borrow for a long trip?
\b. {\sc NN}, {\sc VBP}, {\sc DT} , {\sc NN}, {\sc NN}, {\sc NN}, {\sc PRP}, {\sc MD}, {\sc VB}, {\sc IN}, {\sc DT}, {\sc JJ}, {\sc NN}, {\sc ``.''}
\c. [[{\sc NN}, {\sc VBP}, {\sc DT}] , [{\sc VBP}, {\sc DT}, {\sc NN}], \ldots ]

As an alternative, in this implemented example we assign each symbol in the palette a permutation instead of a vector: $\{\Pi_{NN}, \Pi_{VBP}, ...\}$, and encode the entire sequence through permuting a specific vector, $\bar{v}_{\mathrm{labelsequence}}$, generated to be a place holder for label sequencing and identical for all represented utterances. Each triple is then represented by taking the constant vector $\bar{v}_{\mathrm{labelsequence}}$ and passing it through the label permutations for the labels of the sequence. All these resulting subsequence vectors are then added into the representation for the utterance.  Equation \ref{eq:tripleper} shows how this procedure encodes a sequence [{\sc vbp}, {\sc dt}, {\sc nn}].

 \begin{eqnarray}\label{eq:tripleper}
    \Pi_{\mathrm{nn}}(\Pi_{\mathrm{dt}}(\Pi_{\mathrm{vbp}}( \bar{v}_{\mathrm{labelsequence}}))) 
 \end{eqnarray}

In this way every sentence in the experimental set has the set of category label triples encodings added to its vector as shown in Equation \ref{eq:possemlex}.

 \begin{equation}\label{eq:possemlex}
    \bar{U}_{sum} = \bar{U}_{roles} + \sum_{labeltriples}  \Pi_{\mathrm{label_1}}(\Pi_{\mathrm{label_2}}(\Pi_{\mathrm{label_3}}( \bar{v}_{\mathrm{labelsequence}}))) 
 \end{equation}

\begin{figure}\label{sumneighbours}
\begin{tabular}{c|cl}
\hline
Sample & cosine & utterance \\
\hline
  \ref{samplesamplea}   &  0.54   & meanwhile, 500 miles away from the hurricane in West Texas.  \\
  \ref{samplesampleb}   &  0.67   & Lenny Bruce is not afraid. \\
  \ref{samplesamplec}   &  0.58   & i will always be afraid of hurricanes  \\
  \ref{samplesampled}   &  0.55   & I said it before and I'll say it again... people died in this hurricane.  \\
\end{tabular}
\caption{Closest neighbours to probe sentences in Example \ref{sampleprobes} using lexical, constructional, sequence tags, and semantic roles}
\end{figure}

Notable here is that Example~\ref{samplesampleb}---the one-word utterance ``afraid''---does not benefit from added sophisticated processing, for obvious reasons. The more complex Example~\ref{samplesamplec}---``I am afraid of the hurricane''---increases its similarity to its lexically closest neighbour. Example~\ref{samplesampled}---``I said I am afraid of the hurricane''---finds other utterances where the author utters opinions about hurricanes. In this way more complex utterances will find other utterances with similar characteristics.

\section{Randomness and Noise}
A high-dimensional space where observed items are represented by patterns, rather than by one assigned dimension for each observation---a ``localist'' model---allows for a much larger feature space to be embedded in a given dimensionality $d$. For the purposes given in this paper, the number of potential features---the size of the lexicon and the combined size of all potentially interesting constructions---does not occasion more than linear growth for the system in its entirety. In our implemented example, a 2,000-dimensional space will allow the representation of an entire vocabulary, constructional items, sequence labels, and semantic roles, potentially including cooccurrence statistics. The choice of $d$ determines the capacity of the space. As can be expected, a larger dimensionality allows greater capacity: a 100-dimensional space can store less information, i.e. fewer distinct features, for each state than a 2,000-dimensional does. If we wish to aggregate $N$ (near)-orthogonal features by addition into a state vector, their relative cosine distance to that resulting state vector will be $\sqrt(1/N)$. The expected size of $N$ determines how large $d$ must be chosen to be to ensure that the cosine is at a safe margin from the noise threshold occasioned by the randomisation procedure. If a state vector is expected to hold on the order of 100 unweighted feature vectors, the a resulting relative cosine between each feature vector and the aggregated state vector will be 0.1 on average. In a 1,000-dimensional space, this is about three to four times the noise threshold; in a 2,000-dimensional space about five to six times from the noise threshold. 

\begin{figure*}
\input{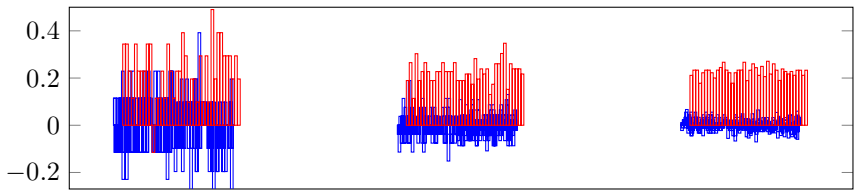}
\caption{Cosine of feature vectors to state vector compared to random unrelated vectors in 100, 500, and 2,000 dimensions
}\label{dimensionalitynoisecomparison}
\end{figure*}

In Figure~\ref{dimensionalitynoisecomparison}, illustrates for four different values of $d$ how a state vector with 20 aggregated random features can retrieve the component features (red bars) compared to random vectors (blue bars). With increasing $d$, the risk of random noise decreases and the accuracy in finding an encoded feature from a state vector increases to near certainty.

\section{Characteristics of high-dimensional spaces} \label{sec:intrinsic}
Very high-dimensional spaces lead human intuitions astray. Navigating in many dimensions is very different from the 2-D and 3-D intuitions we as humans use to understand spatial relationships. We are used to the idea that if $A$ is close to $B$ and $B$ is close
to $C$, then $A$ cannot be very far from $C$.  This intuition fails us
totally in a space of thousands of dimensions if we think in terms of
the ``territory'' (number of points) within a given distance
from point $A$.  Doubling the distance in two-dimensional space
quadruples the territory, whereas in high-dimensional space it can
increase billion-fold.  For example with 2,000-bit vectors a mere
billionth of them are within 865 bits of any vector $A$, but nearly
all are within 1,135 bits. 

This is true of high-dimensional spaces in general, not just binary ones, and it agrees also with the nature of semantic spaces: the words {\it man} and {\it lake}
are far apart, entirely different in meaning, but {\it man} is close to {\it             fisherman} which is close to {\it fish} which is close to {\it                
lake}, and {\it man} is close to {\it plumber} which is close to {\it          
water} which is close to {\it lake}. This can be understood through interpreting the notion of territory above as governing {\em semantic horizon}. Any two linguistic items of notable nearness to each other are related. If their distance is beyond some horizon they are not related. In a high-dimensional space, even a slight divergence from orthogonality, say a cosine of 0.25, is notable. This means that two vectors with a relative angle of $80\degree$ or so are interestingly related, but does not in any way entail that this relatedness should be transitive. 

This has important consequences for the understanding of how feature sets can be aggregated and used in high-dimensional spaces. The sum vectors $\vec{U}$ used in the previous section show that a representation for an utterance encodes every constituent feature in a compact and habitable way. The sum vector retains all the features it has aggregated. The above utterances in Example \ref{sampleprobes} show how they can be used to retrieve other utterances with similar combinations of features. This allows e.g. the search for utterances in a collection by constructing probe utterances which combine features of interest, or by direct query using feature vectors directly. This does not inherently make any claims of how the various features should be kept separate or weighted---this is something that further processing models and classifiers can address given the general representation. 

However, a centroid should not be understood as a generalisation of the features. Adding together e.g. all colour words into a centroid, all names of months into a centroid, or a set of expletives and lewd terms into a centroid does not yield a representation of colourfulness, of months, or of profanity: a vector sum does not in itself provide a generalisation of the component features, but a representation of their combination. This should be kept in mind when similarities are computed.

\section{Suggested work flow}
This paper has presented a model based on high-dimensional computing, in which lexical vector space models can be extended to include sequential, constructional, and more elaborate linguistic items by explicit vector manipulation. This enables the encoding of entire utterances in a vector space similar to what is used for simple lexical items and thus can be integrated with downstream classifiers built to handle lexical representations, unifiying sophisticated processing models with more sophisticated feature sets than previously have been used. The model is {\em explicit}, in that the items under consideration are preserved in the representation and are retrievable from it, which allows for hypothesis-based experimentation. It allows for the inclusion of other vector representations, e.g. from pretrained vocabulary models. It invites joint experimentation with linguistic items of arbitrary sophistication.

The relative weighting of the various features and feature sets encoded in a representation using the above methods will determine how similarity is understood, but there is a general argument about how multiple features can and should be used in a high-dimensional space which we discussed in Section~\ref{sec:intrinsic}. One notable characteristic of this model is that the joint representation of all the above features: lexical items, constructional items, semantic roles, and sequence triples can be used for the features together, or with only some subset of them. If the semantic space built from the joint vectors $\bar{U}_{sum}$ is probed with items encoded with only some of the represented features, the other features will add some level of noise, but not stand in the way of experimentation. Thus, probe utterances such as the ones given above in Example~\ref{sampleprobes} could be encoded for sequential triples alone, to retrieve utterances that have similar constituent structure, with no attention paid to what the lexical content of the utterances are, without need of reencoding the entire collection of items. This versatility of the representation allows for a generous encoding of features to be used in more targeted experimentation as the hypothesis space of a research task becomes more distinct. 


\section{Acknowledgments}
Jussi Karlgren's work was done as a visiting scholar at the Department of Linguistics at Stanford University, supported by a generous {\sc Vinnmer} Marie Curie grant from {\sc Vinnova}, the Swedish Governmental Agency for Innovation Systems. Pentti Kanerva's work was supported by Intel Strategic Research Alliance program on Neuromorphic Architectures for Mainstream Computing and by
NSF 16-526: Energy-Efficient Computing: from Devices to Architectures
(E2CDA), a joint initiative between NSF and SRC.

\bibliographystyle{chicago-nle}
\bibliography{hyperdimensionalutterance}

\label{lastpage}
\end{document}